# A parameterised model for link prediction using node centrality and similarity measure based on graph embedding

Authors: Haohui Lu and Shahadat Uddin

**Affiliation details**

Mr Haohui Lu

School of Project Management

Faculty of Engineering, The University of Sydney, Australia, Level 2, 21 Ross Street, Forest Lodge, NSW 2037

Email: haohui.lu@sydney.edu.au

Dr Shahadat Uddin (Corresponding author)

School of Project Management

Faculty of Engineering, The University of Sydney, Australia, Level 2, 21 Ross Street, Forest Lodge, NSW 2037

Email: shahadat.uddin@sydney.edu.au

**Contribution statement:**

H.L.: Conceptualisation, Data analysis, Research design and Writing; S.U.: Conceptualisation, Research design, Supervision and Writing

**Declarations**

**Conflict of Interests:** The authors declare that they do not have any conflict of interest.

**Funding:** This research did not receive any specific grant from funding agencies in the public, commercial, or not-for-profit sectors.

**Data Availability Statement:**

The datasets generated during and/or analysed during the current study are available from the corresponding author on reasonable request.

# A parameterised model for link prediction using node centrality and similarity measure based on graph embedding

**Abstract**

Link prediction is a key aspect of graph machine learning, with applications as diverse as disease prediction, social network recommendations, and drug discovery. It involves predicting new links that may form between network nodes. Despite the clear importance of link prediction, existing models have significant shortcomings. Graph Convolutional Networks, for instance, have been proven to be highly efficient for link prediction on a variety of datasets. However, they encounter severe limitations when applied to short-path networks and ego networks, resulting in poor performance. This presents a critical problem space that this work aims to address. In this paper, we present the Node Centrality and Similarity Based Parameterised Model (NCSM), a novel method for link prediction tasks. NCSM uniquely integrates node centrality and similarity measures as edge features in a customised Graph Neural Network (GNN) layer, effectively leveraging the topological information of large networks. This model represents the first parameterised GNN-based link prediction model that considers topological information. The proposed model was evaluated on five benchmark graph datasets, each comprising thousands of nodes and edges. Experimental results highlight NCSM's superiority over existing state-of-the-art models like Graph Convolutional Networks and Variational Graph Autoencoder, as it outperforms them across various metrics and datasets. This exceptional performance can be attributed to NCSM's innovative integration of node centrality, similarity measures, and its efficient use of topological information.

**Highlights:**

- Introduced Node Centrality and Similarity Based Parameterised Model (NCSM) for link prediction.
- NCSM integrates node centrality and similarity as edge features in a custom Graph Neural Network.
- Evaluated on five benchmark datasets, outperforming state-of-the-art models like GCNs and Variational Graph Autoencoder.
- NCSM's performance is enhanced by innovative integration of node measures and efficient use of network topology.

# 1. Introduction

A graph comprising nodes and links is a typical but distinctive data structure we often encounter [1]. The design of methods for tasks with graph datasets has proven effective in many real-world applications as machine learning has advanced [2]. Link prediction, which attempts to predict the probability of edge existence between node pairs, is one of the critical challenges in the graph domain [3]. Given the prevalence of graph-structured data, link prediction offers a wide range of applications, including knowledge graph completion [4], disease prediction [5], drug discovery [6], criminal networks [7], and movie recommendation [8].

Significant work in link prediction methods, known as similarity-based approaches, is based on heuristics that anticipate whether nodes will have a link. For example, the Jaccard similarity [9], common neighbour [10], preferential attachment [11], and resource allocation [12] are the commonly used similarity-based methods. While these heuristics may predict links with good performance in some graphs, these approaches rely entirely on the local topological features of a network. When employing these heuristics, it is necessary to manually select different heuristics for distinct graphs based on prior assumptions or extensive experience [13].

Alternatively, the field of machine learning has advanced rapidly in recent years, leading to the development of several models in the graph domain [14]. Graph embedding, or embedding-based methods for link prediction, express a portion of the entire graph in a low-dimensional vector space. Thus, the embedding vector of a link is generated based on the embedding of nodes. Afterwards, a classifier is used to identify whether or not the connection exists [14]. Many embedding-based have been proposed and implemented effectively, including random walk-based [15,16] and spectral clustering [17]. Variational Graph AutoEncoders (VGAE) [18], another graph embedding technique that leverages neighbours' information while creating latent representations for nodes using Graph Convolutional Networks (GCN) [19], has demonstrated high performance on diverse datasets for link prediction. VGAE and other graph embedding approaches use neighbourhood aggregation and message passing to infer node properties. However, these techniques may underperform in short-path or ego networks [1,20]. Short-path networks have fewer hops between nodes, which reduces the distinction between local and global attributes, potentially causing the loss of unique node features during aggregation [20]. Further, traditional methods may overlook ego networks' unique structure and centrality properties formed by a central node and its neighbours [1]. As a result, modifications that consider these unique network characteristics could enhance performance.

Given the shortcomings in the link prediction literature, we developed the following research questions:

1. Can we develop a link prediction model that uses node centrality and similarity measurements to make more accurate predictions?
2. Can a Graph Neural Network's link prediction capability be enhanced by incorporating a custom layer to integrate graph topological information?

3. How does the proposed model compare to cutting-edge algorithms across various datasets, including those of a larger scale?

To answer these questions, we introduce a novel embedding-based approach for link prediction that utilises the topological information of the graph, a unique feature that sets it apart from existing methods. We present the Node Centrality and Similarity Measure-Based Parameterised Model (NCSM). Unlike other models, NCSM predicts potential future connections between nodes by employing a unique combination of centrality and similarity measurements from the graph. Its parameterised nature further sets our model apart, allowing users to adjust the prominence of node centrality and similarity measures based on diverse datasets. The proposed NCSM comprises a topological information calculation module and a Graph Neural Network (GNN) module. These modules offer the dual benefit of extracting valuable topological information from the data and integrating this information directly into the graph learning process. We also introduce a custom layer in the GNN module, specifically designed to incorporate graph topological information.

To summarise, this article offers the following novel contributions:

1. It introduces an innovative framework using an embedding-based link prediction method that can incorporate node centrality and similarity measures in a parameterised model as edge features, allowing for task-specific fine-tuning in link prediction.
2. It develops a unique custom layer in a GNN designed to incorporate edge features and transfer topological information, providing a more sophisticated handling of graph topological data.
3. It conducts comprehensive experiments on several datasets to validate the performance of the proposed framework, demonstrating its capability to handle large-scale data and produce better results than existing models.

The rest of the paper is organised as follows. Section 2 outlines the preliminaries, Section 3 presents an overview of related link prediction studies. Section 4 illustrates the construction of the parameterised model. Section 5 tests the proposed methods on five datasets. Finally, Section 6 presents the conclusion.

## 2. Preliminaries

This section contains the foundational knowledge used in this study. We describe the various node centralities and similarity measures utilised in our work, as well as the basic concept of GNNs.

### 2.1 Node Centrality

Due to the highly heterogeneous structure of complex networks, certain nodes may be considered more important than others [21]. Therefore, it is essential to analyse each node's importance within the network. Node centralities are the most commonly used methods for measuring the significance of nodes in a network. These methods assign nodes a score based on the importance of their topological information relative to certain network factors [1]. To effectively capture the features of nodes in the network, we employ eigenvector centrality and evaluate degree centrality and betweenness centrality in the ablation study section.

**Degree centrality (DC)**, defined by the number of links associated with each node, is the most basic measure of centrality [22]. We use a normalised form in this study, with the formula given by

$$C_D(u) = \frac{d_u}{n-1} \tag{1}$$

Where $d_u$ is the degree of a node $u$ and $n$ is the size of a network.

**Eigenvector centrality (EC)** is determined by each node's neighbourhood. We can infer that a node is significant if it is connected to other significant nodes [23]. Therefore, if 'importance' is measured by vector $x$, we define the importance of node $u$ as

$$x_u = \frac{1}{\lambda} \sum_{v=1}^{n} A_{u,v} x_v, \tag{2}$$

Where $\lambda$ is a constant and not equal to zero. In matrix form:

$$Ax = \lambda x \tag{3}$$

Where $A$ is the adjacency matrix.

**Betweenness centrality (BC)** quantifies the number of shortest paths passing through a specific node. A node is considered significant if it lies on many of the shortest pathways between other nodes[22]. The betweenness centrality of the node $u$ is

$$C_b(u) = \sum_{s \neq u \neq t} \frac{\sigma_{st}(u)}{\sigma_{st}} \tag{4}$$

Where $\sigma_{st}(u)$ is the number of shortest paths between nodes $s$ and nodes $t$ that include nodes $u$, $\sigma_{st}$ is the shortest path between nodes $s$ and nodes $t$ other than nodes $u$.

## 2.2 Similarity Measures

Similarity measures compute the score for each pair of nodes $u$ and nodes $v$. The score between any pair of unconnected nodes can be calculated using various structural and topological network parameters. A high value of a similarity measure an reflect the expected link between them. Our study used the Resource Allocation index and compared Jaccard similarity and Adamic Adar index in the ablation study section.

**Jaccard similarity (JA)** evaluates elements from two sets to ascertain which are common and which are unique [24]. Jaccard index of nodes $u$ and $v$ is defined as:

$$S_{Jaccard}(u,v) = \frac{|N_u \cap N_v|}{|N_u \cup N_v|} \tag{5}$$

Where $N_u$ denotes the set of neighbours of nodes $u$.

**Adamic Adar index (AA)** predicts graph links based on the number of shared links between two nodes [24]. The formula for Adamic Adar is:

$$S_{Adamic\ Adar}(u, v) = \sum_{w \in N_u \cap N_v} \frac{1}{\log(|N_w|)} \tag{6}$$

Where $N_u$ denotes the set of neighbours of nodes $u$. This index results in zero-division for nodes connected only by self-loops. It is intended to be utilised when there are no self-loops.

**Resource Allocation index (RA)** is a measure that calculates the closeness of nodes based on their shared neighbours [12]. This index is defined as:

$$S_{Resource\ Allocation}(u, v) = \sum_{w \in N_u \cap N_v} \frac{1}{|N_w|} \tag{7}$$

Where $N_u$ denotes the set of neighbours of nodes $u$.

## 2.3 Graph neural networks

In recent years, research effort in graph machine learning has grown substantially. Unlike images and sequence data, graphs lack a well-defined node order. Therefore, an architecture that is indifferent to order is required. GNNs are the principal group of models driving this change. GNNs are neural networks that model graphs through a sequence of local message aggregation and propagation stages. GNNs generate vector representations of graph components that encompass information about both graph network topology and node features information [25]. The basic idea of GNNs is that each message-passing layer produces an update to the embeddings of each node in the graph. The propagation at layer $l$ of Graph $G$ is given a pair of nodes $u$ and $v$ is defined as:

$$h_u^{(l)} = UPDATE\left(AGG\left(MSG\left(h_u^{(l-1)}, h_v^{(l-1)}\right) \middle| v \in \mathcal{N}_u\right), h_u^{(l-1)}\right) \tag{8}$$

Where $UPDATE$ denotes a function to update node embeddings, $AGG$ is the aggregation function, $MSG$ is the message passing function. $h_u^{(l-1)}$ represents node embedding for node $u$ at the previous $l-1$ layer. $\mathcal{N}_u$ is set of neighbours of node $u$.

The selection of message passing, aggregation and update functions varies amongst GNN layers. GCN [19], GraphSAGE [26], and Graph Attention Networks (GAT) [27] are some examples of such selection variances. GraphSAGE is an inductive GNN model for generating node embeddings. Instead of using all adjacency matrix information among nodes, it produces aggregator functions that can generate the new node embedding, given its features and neighbourhood information, without retraining the entire model [26]. Due to the flexibility and variability of its architecture, we use this model as our base model. Its message-passing layer is defined as:

$$h_{\mathcal{N}(v)}^{(l)} = AGG\left(h_u^{(l-1)}, \forall u \in \mathcal{N}(v)\right) \tag{9}$$

where $\mathcal{N}(v)$ is the neighbourhood of node $v$, $AGGREGATE$ is the aggregator, which can be mean, LSTM or pooling. Then,

$$h_v^{(l)} = \sigma \left( W^{(l)} \cdot CONCAT \left( h_v^{(l-1)}, h_{\mathcal{N}(v)}^{(l)} \right) \right) \tag{10}$$

where $W^{(l)}$ are the trainable weight matrices, $\sigma(\cdot)$ is a non-linear activation function, such as ReLU.

## 3. Related Work

Link prediction is a significant research topic in network analysis. Numerous link prediction approaches have been proposed in the literature, which falls under three categories: heuristic approaches, embedding methods, and deep learning methods.

### 3.1 Heuristic approaches

Heuristic methods compute the structural similarity between two nodes. These similarities are considered when determining the probability of link formation between nodes. Nodes with high similarity often establish a relationship in the future [28]. Local similarity-based approaches commonly include the Jaccard index [9], common neighbour [10], preferential attachment [11], and resource allocation [12]. Global similarity-based approaches, on the other hand, frequently employ the Katz index [29] and SimRank [30]. With various heuristic techniques designed to handle different graphs, selecting an appropriate heuristic method poses a significant challenge. While heuristic methods are straightforward, computationally economical and mainly require only network structure, they can fall short in accuracy and generalisability and fail to consider node or edge features [31].

### 3.2 Embedding methods

Embedding methods for link prediction involve transforming nodes into vector representations such that the similarity of these vectors may effectively predict the likelihood of a link between nodes in a network. Matrix factorisation (MF) [32] and the stochastic block model [33] for link prediction are methods that predict links using latent node features derived from the network structure. DeepWalk [15] and node2vec [16] are embedding-based methods that employ random walks to generate node embeddings. These embeddings encode latent features and allow pairwise comparison for link prediction tasks. Embedding techniques excel at capturing complex structural patterns and are adaptable to a wide range of tasks. However, they depend on network structure, are computationally costly, and may be challenging to interpret [34].

### 3.3 Graph-based deep learning method

Recent graph-based deep learning methods that are comparable to our proposed model include GCN [19], Graph Variational Autoencoder [18] and GraphSAGE [26], which provide end-to-end graph embedding. Graph Neural Networks and their variants are eventually machine learning architectures that can handle graph data. They are permutation-equivariant models that function by passing data messages through graph structure. A GNN is a highly adaptable and flexible architecture with numerous conceivable augmentations. Lately, researchers developed more GNN variants over time. Ahn and Kim [35] proposed a new Variational Graph Normalized AutoEncoder (VGNAE) that uses L2-normalisation to improve embeddings for isolated nodes. Wu et al. [36] presented a method called MTGC, consisting of several multi-task learnings, each of which learns the information for the link prediction task. Opolka and Lió [37] combined the Bayesian graph convolutional model and Gaussian processes to address over-smoothing and overfitting.

Some studies utilise node features for link prediction. Zhang and Cheng [38] developed a model called SEAL (learning from Subgraphs, Embeddings, and Attributes for Link prediction). They applied GNNs in a non-probabilistic context. Their model is based on node features and hand-crafted node labels that reflect the importance of a node in its neighbourhood. A multi-scale approach was later developed to improve SEAL's performance [39]. Ahmad et al. [40] used a combination of node features (i.e., common neighbour and centrality) to perform similarity-based link prediction. Moreover, several studies have integrated hand-crafted features extracted from networks and supervised machine learning methods for link prediction [41,42]. Deep learning methods on graphs perform better and can handle more complicated features. However, they can overfit, are computationally costly and require large datasets [2]. They also need precise hyperparameter tuning and may be difficult to interpret [43].

Many of the specific elements of previous models are incorporated in the parameterised model described in the following sections. Our model considers graph structure, node features, similarity, and local neighbourhood information when predicting missing links.

## 4. Proposed Methodology

This section details our proposed parameterised model for link prediction, **which utilises** node centrality and similarity measures based on graph embedding.

### 4.1 Motivation

**Numerous** GNN-based link prediction methods have been proposed and proven to have a high performance by leveraging the structure capture capabilities of GNNs [38]. However, for these methods to achieve optimal performance, the dataset must exhibit homophily [44]. Due to the presence of noise, gathering a graph that exhibits homophily is a challenging process in real-world scenarios. Consequently, we consider node centrality to resolve issues with a lack of connections between nodes and similarity measures to overcome feature diversity. We also assign a weight that controls the importance of centrality and similarity for edge features.

## 4.2 Network Architecture

We define an undirected graph $G = (V, A)$, where $V$ is a collection of nodes and $A$ is the adjacency matrix. In this matrix, $A = 1$ indicating the presence of a link between $v_i$ and $v_j$ and $A_{ij} = 0$ indicates the absence of such a link. Fig. 1 illustrates the overall network architecture, which is divided into three key components: parameterised model, GNN and multilayer perceptron predictor.

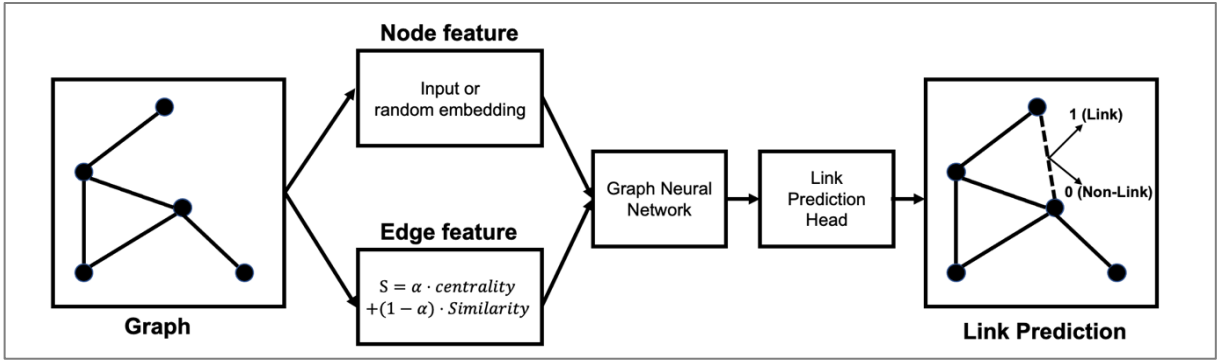

Fig. 1 Overall workflow of the proposed model

### 4.2.1 Parameterised model

Our proposed parameterised model is based on two critical node features: node centrality and similarity between nodes. We have combined three classic node centralities and three similarity measures, as discussed in sections 2.1 and 2.2. The parameterised score for node $u$ is as follows:

$$S = \alpha \cdot centrality + (1 - \alpha) \cdot similarity \tag{11}$$

Where parameter $\alpha$ is a user-defined value between 0 and 1 that governs the weight or importance of centrality and similarity. Centrality can be degree centrality, eigenvector centrality and betweenness centrality. Similarity can be Jaccard similarity, Adamic Adar index and Resource Allocation index. The score from the parameterised model will be used as an edge feature in the GNN model.

### 4.2.2 Custom graph neural network model

GraphSAGE is a powerful architecture for the link prediction task, but it can be improved. The original GraphSAGE model did not consider edge features. In this study, we modify the update equation. To fully exploit the edge feature, the aggregation stage of GraphSAGE is modified as follows:

$$h_v^{(l)} = W_1 \cdot COMB\left(h_v^{(l-1)}, W_3 \cdot S_v\right) + W_2 \cdot AGG\ \left(COMB\left(h_u^{(l-1)}, W_3 \cdot S_u\right), \forall u \in \mathcal{N}(v)\right) \tag{12}$$

Where $W_i$ is a learnable weight matrix, $COMB$ is a combination function that can be sum, mean etc.

### 4.2.3 Link Prediction Head and Evaluation

After deriving the node embeddings from the custom GNN model, we need to use them to perform link prediction tasks. The link prediction head is a binary classification model that determines the probability of an edge existing between two nodes. We need to train this prediction head along with the GNN. As such, we use a deep neural network for link prediction. To evaluate performance, the framework of the confusion matrix can be applied. We use metrics such as True Positive (TP), False Positive (FP), True Negative (TN), and False Negative (FN). The Area Under the Receiver Operating Characteristic curve (AUC), Average Precision (AP), and Hits@20 are used for this study.

The AUC value represents the probability that a randomly chosen missing link between two nodes would be assigned a higher similarity score than a randomly chosen pair of unconnected links [45]. The formula is:

$$AUC = \frac{n' + 0.5n''}{n} \tag{13}$$

Where $n$ is the number of independent comparisons, $n'$ is the number of the missing link having a higher score and $n''$ is the number of comparisons when they have the same score.

A precision-recall curve is summarised by AP as the weighted mean of precisions at each threshold [46], the formula is:

$$AP = \sum_n (R_n - R_{n-1}) P_n \tag{14}$$

Where $P_n$ is the precision and $R_n$ is the recall at the $n^{th}$ threshold.

Hits@20 computes how many components of a rankings ranks vector are present in the top $k$ places [47]. It is defined as follows:

$$Hits@k = \frac{|\{t \in \mathcal{K}_{test} | rank(t) \leq k\}|}{|\mathcal{K}_{test}|} \tag{15}$$

Where $\mathcal{K}_{test}$ is the test set, $rank(t)$ is the ranking of t, $k$ is the top $k$ number. In this study, we select $k = 20$.

Algorithm 1 is a summary of our algorithm.

---
**Algorithm 1: NCSM**

**Input**: Graph $G$, positive edges $\varepsilon_{train}$, random sample negative edges $\bar{\varepsilon}_{train}$, embedding dimension $d$, learning rate $\eta$, total epoch $T$.
**Output:** Link prediction (binary) for each node.
  1: Compute S for all nodes in $G$ by Eq. (11);
  2: **If** node feature exists:
  3:   use node feature $X = \boldsymbol{x}$
  4: **else:**
  5: Random embedding for $\boldsymbol{x}$
  6: **for** *epoch* in range ($T$) **do**
  7:   Get $h_u$ and $h_v$ using Eq. (12)
  8:   Link Prediction = *sigmoid* ($h_u * h_v$)
  9: **end for**
---

## 5. Experiment

### 5.1 Datasets

We use five benchmark datasets, both with and without node features, to test our proposed model, including citation networks: Cora [48], CiteSeer [49], PubMed [50], social networks: Facebook [51]) and drug interaction network: obgl-ddi [52]. We selected these datasets for their broad range of network properties. They span from small graphs to large graphs, with or without node features, and originate from various application fields. Table 1 displays the descriptive statistics for these datasets.

Table 1 Descriptive statistics for network datasets

| Dataset  | Nodes | Edges   | Features | Class |
|----------|-------|---------|----------|-------|
| Cora     | 2708  | 5429    | 1433     | 7     |
| CiteSeer | 3312  | 4660    | 3703     | 6     |
| PubMed   | 19717 | 44338   | 500      | 3     |
| Facebook | 4039  | 88234   | 1283     | 10    |
| ogbl-ddi | 4267  | 1334889 | -        | -     |

## 5.2 Baselines

We evaluate the performance of the proposed model against state-of-the-art link prediction methods. **Matrix Factorisation (MF)** combines latent and optional explicit features in the graph. Various embeddings are allocated to different nodes and are trained end-to-end with the MLP predictor [52]. **Node2Vec** is a technique for mapping nodes to an embedding space. In this study, we use MLP for the downstream link prediction task after embedding. **GCN** is a variant of GNN that applies a convolutional neural network over a graph [19]. **GraphSAGE** is another variant of GNN and a framework for learning inductive representations on large graphs. **Variational Graph Autoencoders (VGAE)** [18] is a graph embedding technique that leverages neighbours' information while creating latent representations for nodes using GCN [19]. **Contrastive Multi-View Representation Learning on Graphs (MVGRL)** is a self-supervised for learning representations by contrasting structural views of graphs [53], and **SEAL** [38] exploits GNNs in a non-probabilistic context using node features and hand-crafted node labels.

## 5.3 Experimental setup

The task in this study is link prediction, which is a binary classification problem. Therefore, the model inputs are node pairs. To create these inputs, we divide our network's positive edges into training, validation and test sets. During training, we also generate an equal number of negative edges by random sampling. For Cora, CiteSeer, PubMed, and Facebook, we remove 5% of the existing edges in the graph as positive samples of the validation set, 10% as positive samples of the test set, and generate an equal number of non-existent edges as negative samples. For ogbl-ddi, we use the default training set, validation set, and test set from the dataset. The validation and test set links are hidden out of the training sets.

We choose our model size to be comparable to the GraphSAGE model in the open graph benchmark [52] and use similar hyperparameters settings. Thus, our model has four layers, each with 256 neurons. We use the Adam optimiser [54] to train our model 500 times with a learning rate of 0.01. We keep the settings from the equivalent studies for other baseline models. We utilise binary cross-entropy loss since link prediction is fundamentally a binary classification task. For the parameterised model setting, we use $\alpha = 0.5$ for initial training.

AUC and AP are chosen as metrics for each experiment, and the method is repeated ten times to produce their averages and standard deviations. In addition, we report Hits@20, one of the key measures on the open graph benchmark leaderboard [52], as a more difficult metric since it requires models to rank positive edges higher than roughly all negative edges.

The graph machine learning methods, node centralities, and similarity measures are implemented in Python with the PyG [55] and Networkx [56] packages, respectively. The experiments are performed on a server with Tesla P100 GPU (16 GB memory).

## 5.4 Results and Discussions

The link prediction problem is challenging due to implicit, latent relationships and many potential link candidates. Traditionally, link prediction has relied heavily on the features of individual nodes. However, these features cannot capture the broader context of the node's position within the graph's overall structure. We improve our understanding of the network's overall structure by including edge features in the model. Despite their apparent simplicity, they provide the model with a deeper, more contextual understanding of the network, resulting in a considerable improvement in prediction performance.

The experimental results, summarised in Tables 2, 3 and 4, reveal that our proposed model, NCSM (RA+EC), outperforms the baseline models across all three performance metrics: AUC, AP, and Hits@20. They demonstrate that by including a parameterised model as an edge feature in GraphSAGE, our model obtains the best results and substantially outperforms the baselines. In particular, NCSM performs well on large datasets (PubMed, Facebook, and obgl-ddi).

Compared to the baseline approaches with the best link prediction performance measures (i.e., the methods underlined in Tables 2, 3, and 4), the proposed method, NCSM, performs better on AUC, AP, and Hits@20. Further, Fig. 2 compares the experimental results between NCSM and the baseline models. It can be seen that our proposed model outperforms the baseline methods in terms of link prediction. Specifically, the increase ranges from 0.7% (in the Facebook dataset) to 11% (in the CiteSeer dataset). This boost in the AUC score signifies that our model is significantly better at distinguishing positive links from negative ones, indicating superior prediction ability. In terms of AP scores (Table 3), our model again outperforms the baselines, with improvements ranging from 0.3% (Cora dataset) to 4.4% (CiteSeer dataset). Finally, we consider Hits@20 (Table 4), which requires the model to rank positive edges much higher than negative ones. NCSM again comes out on top, with improvements ranging from 9% (on the ogbl-ddi dataset) to as much as 15% (on the Citeseer dataset). This substantial increase in the Hits@20 score indicates our model's ability to prioritise and rank true positive links more effectively, which is crucial for real-world applications.

In summary, we attribute the improvements in performance to a variety of factors introduced by adding edge features. The unique advantage of NCSM is its use of node centrality and similarity measurements as edge features, which are often ignored by other GNN-based link prediction algorithms. These features provide topological information that is critical for link prediction since it reveals the network's underlying structure and interactions. As a result, the model may learn more efficient data representations, significantly improving link prediction performance.

Table 2 AUC for link prediction. Bold denotes the best performances for the proposed method. Underline denotes the best-performed baseline. The standard deviation is shown in brackets.

|  | Cora | Citeseer | PubMed | Facebook | obgl-ddi |
| --- | --- | --- | --- | --- | --- |
| MF | 0.5054 (0.0082) | 0.5020 (0.0142) | 0.5529 (0.0024) | 0.7050 (0.0098) | 0.8659 (0.0459) |
| Node2Vec | 0.8668 (0.0024) | 0.8276 (0.0098) | 0.7935 (0.0099) | 0.8691 (0.0027) | 0.9202 (0.0020) |
| GCN | 0.9025 (0.0053) | 0.7147 (0.0140) | 0.9633 (0.0080) | 0.9943 (0.0020) | 0.9982 (0.0050) |
| GraphSAGE | 0.8024 (0.0034) | <u>0.8738 (0.0139)</u> | <u>0.9678 (0.0011)</u> | 0.9929 (0.0040) | <u>0.9993 (0.0020)</u> |
| MVGRL | 0.7507 (0.0363) | 0.6120 (0.0550) | 0.8078 (0.0128) | 0.7983 (0.0300) | 0.8145 (0.0990) |
| VGAE | 0.8868 (0.0400) | 0.8535 (0.0060) | 0.9580 (0.0130) | 0.9866 (0.0040) | 0.9308 (0.0150) |
| SEAL | <u>0.9255 (0.0050)</u> | 0.8582 (0.0440) | 0.9636 (0.0280) | <u>0.9960 (0.0002)</u> | 0.9785 (0.0017) |
| NCSM (RA+EC) | **0.9321 (0.0154)** | **0.9527 (0.0154)** | **0.9740 (0.0176)** | **0.9968 (0.0022)** | **0.9995 (0.0030)** |

Table 3 AP for link prediction. Bold denotes the best performances for the proposed method. Underline denotes the best-performed baseline. The standard deviation is shown in brackets.

|  | Cora | Citeseer | PubMed | Facebook | obgl-ddi |
| --- | --- | --- | --- | --- | --- |
| MF | 0.5036 (0.0045) | 0.5025 (0.0072) | 0.5451 (0.0017) | 0.6646 (0.0057) | 0.8844 (0.0366) |
| Node2Vec | 0.8377 (0.0043) | 0.8179 (0.0085) | 0.7889 (0.0108) | 0.8740 (0.0085) | 0.9188 (0.0017) |
| GCN | 0.9046 (0.0152) | 0.8796 (0.0175) | 0.8814 (0.0111) | 0.9470 (0.0015) | 0.9818 (0.0009) |
| GraphSAGE | 0.8963 (0.0102) | 0.8858 (0.0132) | 0.8984 (0.0141) | 0.9526 (0.0017) | <u>0.9921 (0.0002)</u> |
| MVGRL | 0.7456 (0.0424) | 0.6326 (0.0752) | 0.8352 (0.0448) | 0.8086 (0.2310) | 0.8177 (0.0865) |
| VGAE | 0.9089 (0.0335) | 0.8696 (0.0560) | 0.9370 (0.2131) | 0.9875 (0.0865) | 0.9338 (0.1150) |
| SEAL | <u>0.9090 (0.0750)</u> | <u>0.8982 (0.0423)</u> | <u>0.9399 (0.0780)</u> | <u>0.9559 (0.0012)</u> | 0.9895 (0.0037) |
| NCSM (RA+EC) | **0.9096 (0.0172)** | **0.9382 (0.0180)** | **0.9411 (0.0170)** | **0.9589 (0.0022)** | **0.9955 (0.0003)** |

Table 4 Hits@20 for link prediction. Bold denotes the best performances for the proposed method. Underline denotes the best-performed baseline. The standard deviation is shown in brackets.

|  | Cora | Citeseer | PubMed | Facebook | obgl-ddi |
| --- | --- | --- | --- | --- | --- |
| MF | 0.0495 (0.0156) | 0.0618 (0.0167) | 0.0449 (0.0074) | 0.0406 (0.0059) | 0.1368 (0.0475) |
| Node2Vec | 0.4523 (0.0986) | 0.4808 (0.0118) | 0.3898 (0.0866) | 0.2223 (0.0986) | 0.2326 (0.0209) |
| GCN | 0.4906 (0.0172) | <u>0.5556 (0.0132)</u> | 0.2184 (0.0387) | 0.5389 (0.0214) | 0.3707 (0.0507) |
| GraphSAGE | 0.5354 (0.0296) | 0.5367 (0.0294) | <u>0.3913 (0.0441)</u> | <u>0.4551 (0.0322)</u> | <u>0.5390 (0.0474)</u> |
| MVGRL | 0.1953 (0.0264) | 0.1407 (0.0079) | 0.1419 (0.0085) | 0.1443 (0.0033) | 0.1002 (0.0101) |
| VGAE | 0.4591 (0.0338) | 0.4404 (0.0486) | 0.2373 (0.0161) | 0.3701 (0.0063) | 0.1171 (0.0196) |
| SEAL | <u>0.5135 (0.0226)</u> | 0.4090 (0.0368) | 0.2845 (0.0381) | 0.4089 (0.0570) | 0.3056 (0.0386) |
| NCSM (RA+EC) | **0.6072 (0.0779)** | **0.6625 (0.0647)** | **0.4445 (0.0519)** | **0.5586 (0.0558)** | **0.6500 (0.1166)** |

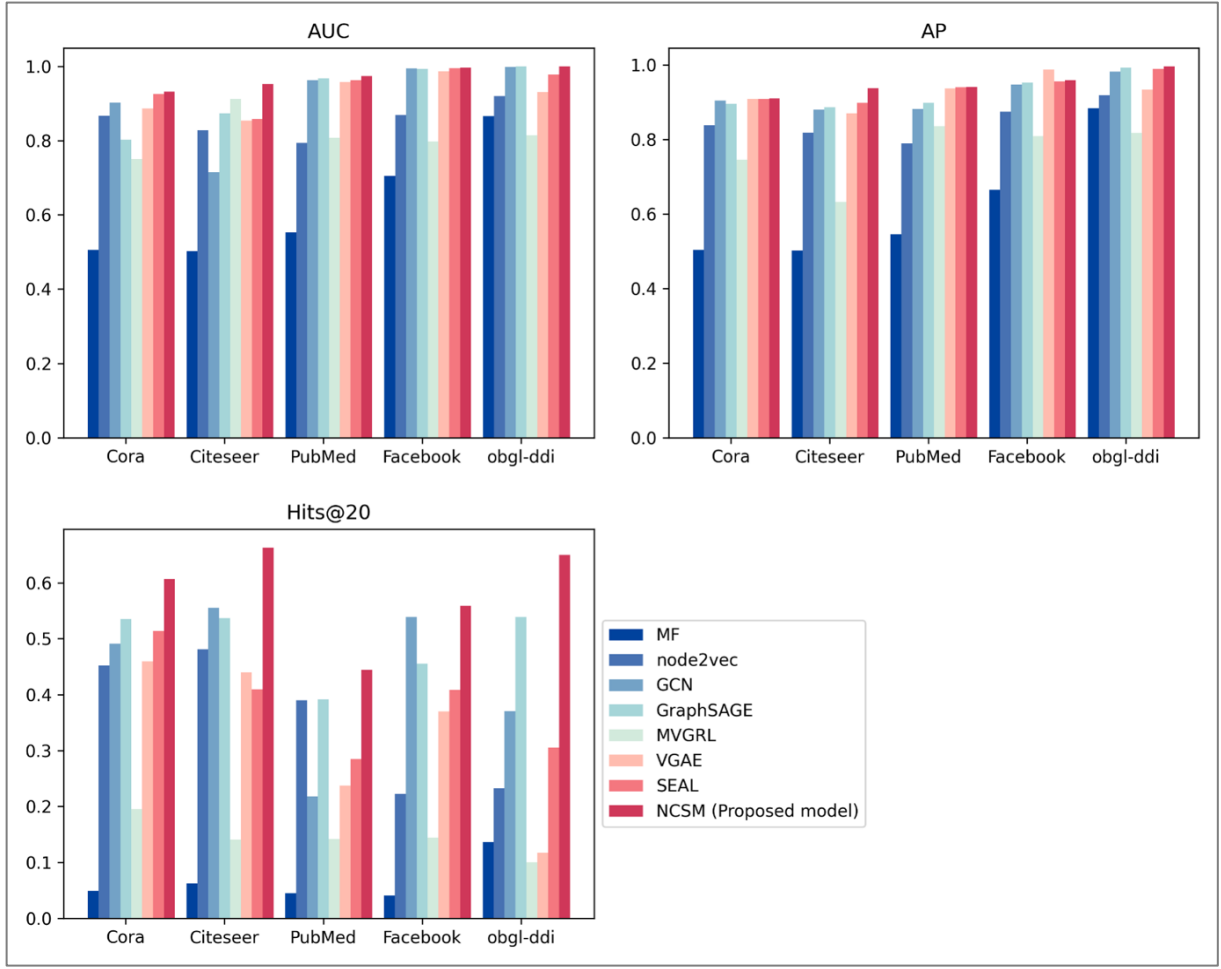

Fig. 2 Comparison of results for proposed NCSM model and baseline models

## 5.5 Ablation Study

We conducted an ablation study to better understand the role of different components of NCSM. By evaluating the performance with different combinations of node centralities and similarity measures (Tables 5, 6, and 7), we discovered that the combination of the Resource Allocation (RA) index for node similarity and Eigenvector Centrality (EC) for node centrality (NCSM (RA+EC)) consistently produced the best results. This suggests that when employed together, the RA index and the EC capture complementary features of the graph structure that contribute to the effectiveness of link prediction.

### 5.5.1 Different Node Centralities and Similarity Measures

As can be seen, the combination of RA and EC achieves the best result among night variants. In addition, adopting the combination of RA and EC is also effective on certain large data sets. As mentioned in Section 3.2, RA, which calculates the closeness of nodes based on their common neighbours, appears to be particularly effective on certain large datasets. This makes intuitive sense since the presence of common neighbours often indicates a higher likelihood of a link forming between two nodes. EC, on the other hand, measures the influence of a node in a network considering both the number and the quality of its connections. This could be particularly useful for identifying potential links involving influential nodes, which tend to form connections with other high-centrality nodes. These two features combine the importance and closeness levels of the nodes. Further, EC is substantially more computationally efficient than BC and does not require similarity value thresholding.

Table 5 AUC for NCSM with different centralities and similarities. Bold denotes the best performances. The standard deviation is shown in brackets.

|  | Cora | Citeseer | PubMed | Facebook | obgl-ddi |
| --- | --- | --- | --- | --- | --- |
| NCSM (JA+DC) | 0.9197 (0.0154) | 0.9402 (0.0175) | 0.9405 (0.0104) | 0.9690 (0.0025) | 0.9953 (0.0005) |
| NCSM (JA+EC) | 0.9027 (0.0139) | 0.9480 (0.0108) | 0.9341 (0.0156) | 0.9700 (0.0020) | 0.9954 (0.0002) |
| NCSM (JA+BC) | 0.9211 (0.0141) | 0.9486 (0.0090) | 0.9354 (0.0174) | 0.9691 (0.0014) | 0.9952 (0.0004) |
| NCSM (AA+DC) | 0.9199 (0.0156) | 0.9433 (0.0100) | 0.9356 (0.0109) | 0.9630 (0.0017) | 0.9951 (0.0005) |
| NCSM (AA+EC) | 0.9222 (0.0185) | 0.9476 (0.0123) | 0.9381 (0.0108) | 0.9700 (0.0017) | 0.9952 (0.0005) |
| NCSM (AA+BC) | 0.9102 (0.0234) | 0.9421 (0.0184) | 0.9399 (0.0133) | 0.9700 (0.0019) | 0.9954 (0.0003) |
| NCSM (RA+DC) | 0.9213 (0.0112) | 0.9463 (0.0098) | 0.9388 (0.0144) | 0.9702 (0.0017) | 0.9953 (0.0004) |
| NCSM (RA+EC) | **0.9321 (0.0154)** | **0.9527 (0.0154)** | **0.9740 (0.0176)** | **0.9968 (0.0022)** | **0.9995 (0.0030)** |
| NCSM (RA+BC) | 0.9220 (0.0182) | 0.9447 (0.0167) | 0.9284 (0.0111) | 0.9705 (0.0019) | 0.9953 (0.0004) |

Table 6 AP for NCSM with different centralities and similarities. Bold denotes the best performances. The standard deviation is shown in brackets.

|  | Cora | Citeseer | PubMed | Facebook | obgl-ddi |
| --- | --- | --- | --- | --- | --- |
| NCSM (JA+DC) | 0.9081 (0.0166) | 0.9306 (0.0172) | 0.9301 (0.0106) | 0.9565 (0.0017) | 0.9953 (0.0004) |
| NCSM (JA+EC) | 0.8931 (0.0117) | 0.9346 (0.0177) | 0.9312 (0.0155) | 0.9568 (0.0015) | 0.9954 (0.0002) |
| NCSM (JA+BC) | 0.9094 (0.0101) | 0.9359 (0.0089) | 0.9329 (0.0171) | 0.9566 (0.0018) | 0.9952 (0.0005) |
| NCSM (AA+DC) | 0.9046 (0.0229) | 0.9312 (0.0162) | 0.9326 (0.0108) | 0.9580 (0.0013) | 0.9952 (0.0004) |
| NCSM (AA+EC) | 0.9034 (0.0348) | 0.9332 (0.0136) | 0.9357 (0.0107) | 0.9564 (0.0029) | 0.9951 (0.0005) |
| NCSM (AA+BC) | 0.8956 (0.0264) | 0.9313 (0.0173) | 0.9366 (0.0128) | 0.9580 (0.0014) | 0.9953 (0.0003) |
| NCSM (RA+DC) | 0.9072 (0.0167) | 0.9350 (0.0104) | 0.9358 (0.0136) | 0.9579 (0.0014) | 0.9953 (0.0004) |
| NCSM (RA+EC) | **0.9096 (0.0172)** | **0.9382 (0.0180)** | **0.9411 (0.0170)** | **0.9589 (0.0022)** | **0.9955 (0.0003)** |
| NCSM (RA+BC) | 0.9011 (0.0306) | 0.9309 (0.0150) | 0.9256 (0.0105) | 0.9568 (0.0021) | 0.9954 (0.0003) |

Table 7 Hits@20 for NCSM with different centralities and similarities. Bold denotes the best performances. The standard deviation is shown in brackets.

|  | **Cora** | **Citeseer** | **PubMed** | **Facebook** | **obgl-ddi** |
| --- | --- | --- | --- | --- | --- |
| NCSM (JA+DC) | 0.5863 (0.0284) | 0.6577 (0.0148) | 0.3791 (0.0433) | 0.3170 (0.0250) | 0.5485 (0.1466) |
| NCSM (JA+EC) | 0.5833 (0.0074) | 0.5503 (0.0453) | 0.3366 (0.0466) | 0.2710 (0.0411) | 0.5012 (0.1256) |
| NCSM (JA+BC) | 0.5939 (0.0432) | 0.5618 (0.0228) | 0.3535 (0.0345) | 0.2834 (0.0420) | 0.5796 (0.1234) |
| NCSM (AA+DC) | 0.5850 (0.1054) | 0.5527 (0.0439) | 0.3325 (0.0593) | 0.2843 (0.0631) | 0.5665 (0.1364) |
| NCSM (AA+EC) | 0.5590 (0.1487) | 0.5495 (0.0399) | 0.3670 (0.0509) | 0.3115 (0.0476) | 0.5375 (0.1633) |
| NCSM (AA+BC) | 0.5723 (0.1181) | 0.5607 (0.0151) | 0.3381 (0.0698) | 0.3184 (0.0295) | 0.5723 (0.0804) |
| NCSM (RA+DC) | 0.5896 (0.0843) | 0.5604 (0.0201) | 0.4389 (0.0409) | 0.3100 (0.0167) | 0.5944 (0.0829) |
| NCSM (RA+EC) | **0.6072 (0.0779)** | **0.6625 (0.0647)** | **0.4445 (0.0519)** | **0.5586 (0.0558)** | **0.6500 (0.1166)** |
| NCSM (RA+BC) | 0.5954 (0.1626) | 0.6543 (0.0222) | 0.4263 (0.0689) | 0.2922 (0.0397) | 0.5989 (0.1575) |

### 5.5.2 Different Parameter

The proposed NCSM model contains a parameter $\alpha$. We report the results obtained by performing the proposed model on various values of $\alpha$ to determine the influence on the obtained values of AUC, AP, and Hits@20. Since the graph size of Cora and Citeseer are relatively small, the results are very similar by using different $\alpha$. Therefore, we focus on the larger graph PubMed, Facebook and ogbl-ddi. As mentioned in section 5.4, the combination of RA and EC outperforms other combinations on large graphs, and we use this combination to test different $\alpha$. Fig. 3 shows a graphical representation of the trend in average AUC, AP and Hits@20 for various values of $\alpha$. There is no significant difference in the average AUC, AP, and Hits@20 depending on the change in the value of $\alpha$. Fig. 3 also indicates that for $\alpha = 0.2$, the smallest average value of AUC, AP, and Hits@20 is attained, whereas, for $\alpha = 0.5$, the highest value is acquired. Even across three large graphs, we could not detect a pattern to determine the value of $\alpha$ will yield the best results. Therefore, it will be worthwhile to conduct a comprehensive analysis of the proposed models on a wide range of datasets with different types of graphs to derive a general inference for selecting the ideal value of $\alpha$. For example, the correlation between $\alpha$ and the average degree of the graph or the average clustering coefficient. As a result, these may be the subject of further research.

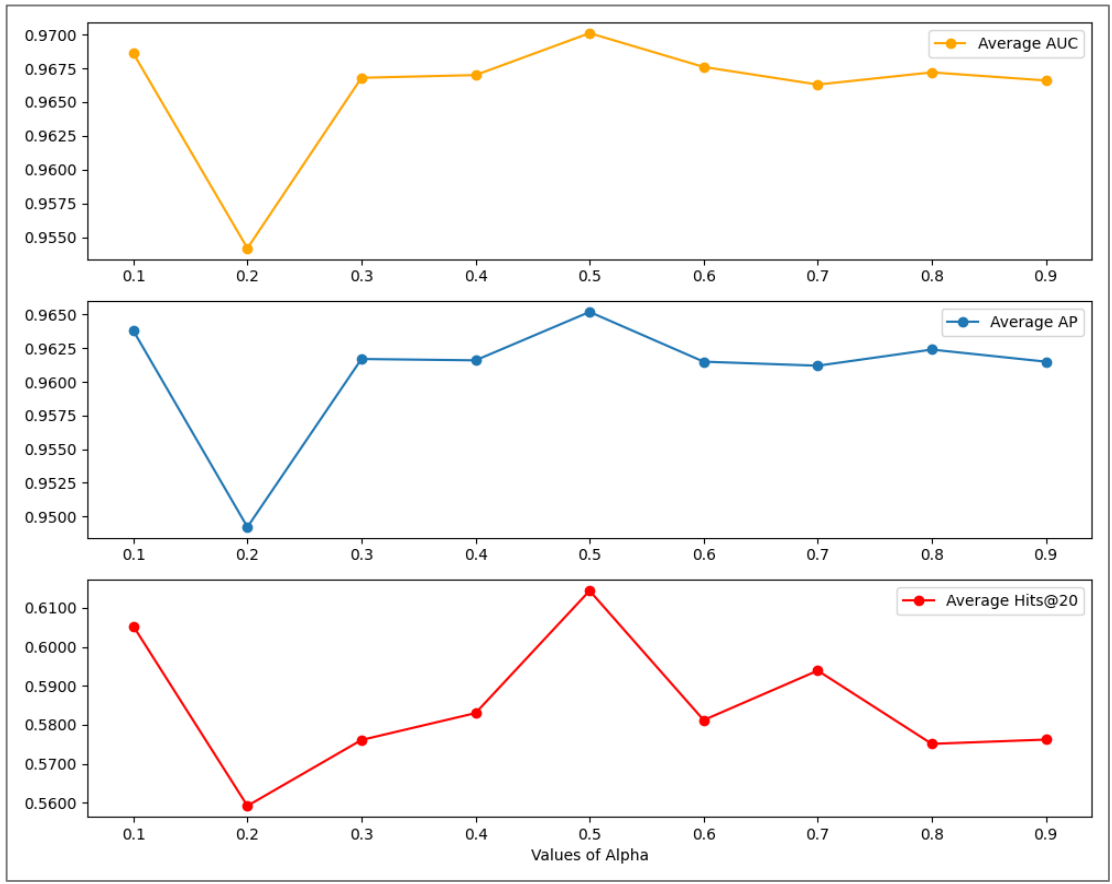

Fig. 3 A graphical representation of the trend in average AUC, AP and Hits@20 for various values of α

## 6 Conclusion

Link prediction has gained popularity in a variety of fields. Existing link prediction methods mainly consider structure information but ignore the topological information of a graph. In this study, we addressed a critical gap in link prediction by proposing the NCSM. This model incorporates node centrality and similarity measures as edge features, using a custom GNN layer, which successfully captures the topological information of large networks. NCSM is the parameterised GNN-based link prediction model that particularly considers topological information. NCSM outperformed existing state-of-the-art models, such as GCNs and VGAE, across five different datasets. The success of NCSM offers a promising new direction for enhancing link prediction by effectively utilising network topological information and holds potential for broader application in various fields.